\documentclass{Interspeech}

\title{Exploring Generative Error Correction for Dysarthric Speech Recognition}

\interspeechcameraready 

\author[affiliation={1}]{Moreno}{La Quatra*}
\author[affiliation={2}]{Alkis}{Koudounas*}
\author[affiliation={1}]{Valerio Mario}{Salerno}
\author[affiliation={3}]{Sabato Marco}{Siniscalchi}

\affiliation{}{Kore University of Enna}{Italy}
\affiliation{}{Politecnico di Torino}{Italy}
\affiliation{}{Università di Palermo}{Italy}

\email{moreno.laquatra@unikore.it, alkis.koudounas@polito.it, valerio.salerno@unikore.it, sabatomarco.siniscalchi@unipa.it}

\keywords{dysarthric speech recognition, speech accessibility, generative error correction}

\usepackage{comment}
\usepackage{booktabs}
\usepackage{multirow}
\usepackage{pifont}
\usepackage{xcolor}
\usepackage[many]{tcolorbox}
\usepackage{footmisc}
\usepackage{capt-of}
\usepackage{capt-of}
\usepackage{cite}

\definecolor{mutedblue}{RGB}{50, 80, 150}

\tcbset{promptbox/.style={
    colback=gray!5, 
    colframe=gray!50, 
    fonttitle=\bfseries\small,
    coltitle=black,
    sharp corners,
    boxrule=0.5pt,
    left=7pt, right=7pt, top=5pt, bottom=5pt,
    width=0.99\linewidth
}}

\definecolor{negative}{RGB}{194,42,37}
\definecolor{positive}{RGB}{107,142,35}
\definecolor{positive}{RGB}{0,100,0}
\definecolor{negative}{RGB}{180,0,0}
\newcommand{\negative}[1]{\textcolor{negative}{#1}}
\newcommand{\positive}[1]{\textcolor{positive}{#1}}

\newcommand{\equalcontrib}{$^{*}$}
\newcommand\blfootnote[1]{%
  \begingroup
  \renewcommand\thefootnote{}\footnote{#1}%
  \addtocounter{footnote}{-1}%
  \endgroup
  }
  
\begin{document}

\maketitle

\begin{abstract}
Despite the remarkable progress in end-to-end Automatic Speech Recognition (ASR) engines, accurately transcribing dysarthric speech remains a major challenge.
In this work, we proposed a two-stage framework for the Speech Accessibility Project Challenge at INTERSPEECH 2025, which combines cutting-edge speech recognition models with LLM-based generative error correction (GER).
We assess different configurations of model scales and training strategies, incorporating specific hypothesis selection to improve transcription accuracy.
Experiments on the Speech Accessibility Project dataset demonstrate the strength of our approach on structured and spontaneous speech, while highlighting challenges in single-word recognition.
Through comprehensive analysis, we provide insights into the complementary roles of acoustic and linguistic modeling in dysarthric speech recognition\footnote{\url{https://github.com/MorenoLaQuatra/GER4Dys}}.
\end{abstract}

\section{Introduction}
\label{sec:introduction}
\blfootnote{\equalcontrib{} Both authors contributed equally to this work.}
Speech recognition technology has become essential in daily life, enabling natural interaction with devices and applications \cite{yang21c_interspeech,koudounas23_interspeech,speech_massive, arch}.
Voice-controlled systems now support tasks ranging from simple commands to complex conversations.
However, these advances have not benefited all users equally~\cite{10095284, 10430478, koudounas2024contrastive}.
Individuals with dysarthric speech face significant challenges when using conventional Automatic Speech Recognition (ASR) systems \cite{leung24_interspeech, dys_asr_study}.
Dysarthric speech, resulting from motor speech disorders, exhibits irregular articulation, atypical prosody, and inconsistent speaking rates.
These speech characteristics are often associated with various medical conditions, as speech patterns serve as important biomarkers for neurological disorders like Parkinson's disease and voice disorders \cite{orozco2014new, koudounas_2024_voice,quatra_2024_exploiting,BDHPD}.
While modern ASR systems have achieved notable improvements through self-supervised learning and large-scale training data \cite{wav2vec2,hubert,whisper,voc2vec}, their error rates on dysarthric speech often exceed 30\% \cite{hasegawa2024community}.

Modern ASR systems using beam-search-based decoding strategies are able to generate multiple possible transcriptions (hypotheses) ranked by confidence scores.
While the top-ranked hypothesis might be incorrect for dysarthric speech, more correct alternatives may appear among lower-ranked candidates (see Section \ref{subs:performance_utter} and Table \ref{tab:ger-example} for more details).
This suggests that ASR models may capture the acoustic information but struggle with the proper ranking of alternatives.
The Speech Accessibility Project (SAP) addresses these challenges by providing a dataset of 140+ hours of dysarthric speech \cite{hasegawa2024community}.
Unlike previous collections, the SAP dataset offers greater scale and detailed speech annotations through contextualized information and verbatim transcriptions.
In this work, we investigate two fundamental questions about dysarthric speech recognition:
\begin{enumerate}
    \item Do general-purpose ASR models possess sufficient acoustic modeling capacity to capture dysarthric speech patterns, even if they struggle with producing correct transcriptions?
    \item Can large language models leverage their linguistic knowledge to identify correct transcriptions by analyzing patterns across multiple ASR hypotheses while ensuring grammatical and contextual coherence?
\end{enumerate}

We propose a two-stage framework that combines ASR with generative error correction.
Our approach first uses ASR models to generate multiple transcription hypotheses, capturing different interpretations of the acoustic input.
A large language model then analyzes these hypotheses collectively, using its understanding of language patterns and context to produce more accurate transcriptions.
\noindent
While previous approaches to dysarthric speech recognition have focused on adapting acoustic models \cite{wang2024enhancing,hsieh24_interspeech}, we also explore error correction through text-to-text mapping using language modeling. 
We evaluate this system as part of the Speech Accessibility Challenge at INTERSPEECH 2025, focusing on developing effective ASR solutions for dysarthric speech.
By combining acoustic and linguistic approaches, our work contributes to making speech technology more accessible for individuals with speech impairments.

\begin{figure*}[ht!]
    \centering
    \includegraphics[width=\textwidth]{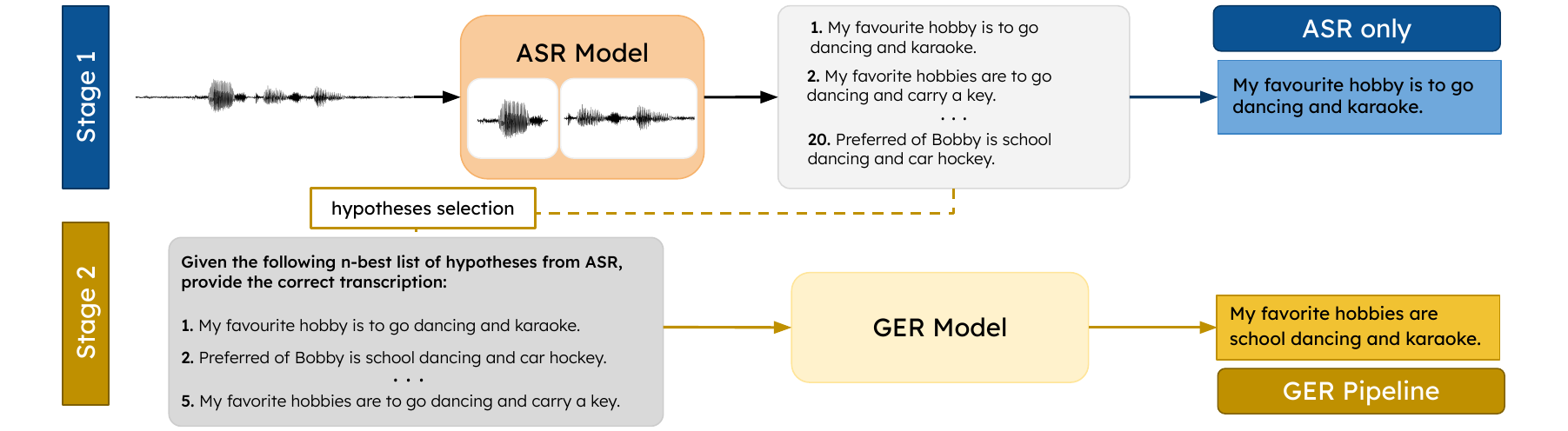}
    \caption{Overview of our two-stage framework for dysarthric speech recognition. 
    Stage 1 uses the ASR model to generate 20-best hypotheses from the input audio. 
    Stage 2 selects diverse hypotheses and employs the GER model to analyze them collectively, producing a refined final transcription.}
    \vspace{-5mm}
    \label{fig:SAPC_pipeline}
\end{figure*}

\section{Methodology}
\label{sec:methodology}

Our work explores (i) whether general ASR systems can capture useful information about dysarthric speech in their N-best hypotheses, and (ii) if language models can effectively distill this information.
To investigate these research questions, we develop a two-stage framework that combines ASR with Generative Error Correction (GER) (Figure \ref{fig:SAPC_pipeline}).
Our framework separates the acoustic processing from linguistic error correction, allowing us to analyze the contribution of each component.

\subsection{ASR Stage}
The first stage of our framework uses Whisper \cite{whisper} for speech recognition, implemented within the WhisperX framework \cite{bain23_interspeech} for efficient processing of long-form audio.
We investigate two distinct approaches to handling dysarthric speech: zero-shot evaluation and fine-tuning.
The zero-shot approach tests how well a general-purpose ASR model handles dysarthric speech without any adaptation.
For adaptation, we fine-tune the entire model to specialize it for dysarthric speech recognition.
For each audio input, we generate N transcription hypotheses through beam search.
This N-best generation allows to capture different interpretations of unclear or challenging speech segments.
For long recordings, WhisperX handles segmentation using integrated voice activity detection \cite{vad,pyannote}.
When processing these multi-segment recordings (i.e., utterances longer than 30 seconds), we maintain consistency by concatenating corresponding ranked hypotheses across segments (e.g., all top-1 hypotheses are concatenated to form the complete top-1 hypothesis).

\vspace{1mm}
\noindent
\textbf{Hypothesis Selection}
To ensure meaningful variation while maintaining computational efficiency, we implement a diversity-based selection algorithm that chooses 5 hypotheses from the initial pool of 20.
It follows these steps:
\begin{enumerate}
    \item Retain the top-scoring hypothesis to preserve the highest confidence transcription
    \item Calculate normalized edit distances between all remaining hypotheses
    \item Iteratively select hypotheses that maximize the minimum distance to previously selected ones
\end{enumerate}
This approach is designed to select hypotheses that represent genuinely different interpretations rather than minor variations of the same transcription.

\subsection{Generative Error Correction}
\begin{figure}[h]
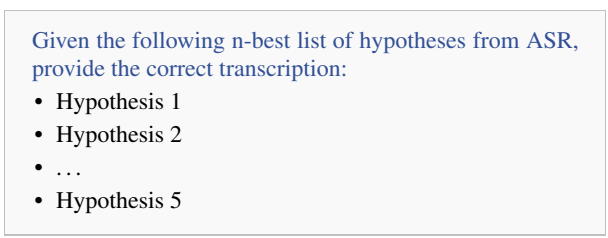

    \centering
    \begin{tcolorbox}[promptbox]
        \textcolor{mutedblue}{Given the following n-best list of hypotheses from ASR, provide the correct transcription:}
        \begin{itemize}
            \item Hypothesis 1
            \item Hypothesis 2
            \item \dots
            \item Hypothesis 5
        \end{itemize}
    \end{tcolorbox}
    \caption{GER model prompt.}
    \label{fig:asr_prompt}
\end{figure}

The second stage of our framework uses a sequence-to-sequence model to analyze the selected hypotheses collectively and generate a refined transcription.
We employ the FlanT5 model \cite{flanT5} which has demonstrated strong capabilities in error correction tasks \cite{flanec}.
Like other sequence-to-sequence models, FlanT5 can process variable-length input sequences and generate coherent text while maintaining semantic meaning.
The model receives the hypotheses through a structured prompt as shown in Figure \ref{fig:asr_prompt}.

\noindent
We train the GER model using LoRA \cite{LORA} to maintain efficiency while specializing in transcription correction.
This approach allows the model to leverage patterns across hypotheses and its own language understanding to resolve ambiguities in the ASR output.
The GER stage serves multiple purposes: (i) analyzes patterns across multiple hypotheses to identify consistent elements, (ii) leverages linguistic knowledge to correct grammatical errors and resolve ambiguities, and (iii) produces a final transcription that balances acoustic evidence with linguistic plausibility.

\vspace{1mm}
\noindent
\textbf{Training Approach}
We train our framework in two distinct phases.
First, we either use the ASR model in zero-shot mode or fine-tune it on our dataset.
Second, we train the GER model using pairs of N-best hypotheses and ground truth transcriptions from the training set.
This two-phase approach allows each component to specialize in its role while maintaining complementary capabilities.

\begin{figure*}[t!]
\begin{minipage}[t]{0.65\textwidth}
\centering
\captionof{table}{Performance comparison across ASR and GER configurations.}
\label{tab:asr-performance}
\begin{tabular}{@{}ccc|cc|cc|cc@{}}
\toprule
\multirow{2}{*}{\shortstack{ASR\\Model}} & \multirow{2}{*}{FT} & \multirow{2}{*}{GER} & \multicolumn{2}{c|}{Dev} & \multicolumn{2}{c|}{\textsc{test-1}} & \multicolumn{2}{c}{\textsc{test-2}} \\
 &  &  & WER & SemScore & WER & SemScore & WER & SemScore \\ \midrule
Lv3 & \negative{\ding{55}} & \negative{\ding{55}} & 11.60 & 83.91 & 11.39 & 84.82 & 14.49 & 78.83 \\
Lv3 & \negative{\ding{55}} & 3B & 7.91 & 89.42 & 10.99 & 86.45 & 13.87 & 79.88 \\
Lv3 & \negative{\ding{55}} & 11B & 7.34 & 90.32 & \textbf{10.63} & 86.96 & 13.64 & 80.4 \\
Lv2 & \positive{\checkmark} & \negative{\ding{55}} & 7.17 & 91.94 & 10.90 & 87.24 & 13.04 & 81.89 \\
Lv2 & \positive{\checkmark} & 3B & \textbf{6.40} & \textbf{92.47} & 10.67 & \textbf{87.51} & \textbf{12.89} & \textbf{82.16} \\ \bottomrule
\end{tabular}
\end{minipage}
\hfill
\begin{minipage}[t]{0.33\textwidth}
\centering
\captionof{table}{Ablation study on N-best list size.}
\label{tab:ablation_n_best}
\begin{tabular}{@{}cc|cc@{}}
\toprule
\multirow{2}{*}{\shortstack{ASR\\Model}} & \multirow{2}{*}{N} & \multicolumn{2}{c}{Dev} \\
 &  & WER & SemScore \\ \midrule
Lv3 & 1 & 12.37 & 83.07 \\
Lv3 & 5 & 11.71 & 83.79 \\
Lv3 & 10 & 11.62 & 83.89 \\
Lv3 & 20 & \textbf{11.60} & \textbf{83.91} \\ \bottomrule
\end{tabular}
\end{minipage}
\vspace{-5mm}
\end{figure*}

\section{Experimental Setup}
\label{sec:experiments}

Our experimental evaluation primarily uses the Speech Accessibility Project Challenge (SAPC) dataset \cite{hasegawa2024community}, containing 105.76 hours of dysarthric speech in the training set and 39.56 hours in the development set.
The dataset consists of multiple utterance types: Digital Assistant Commands (DAC), Sentences from Novels (SN), Spontaneous Speech (SS), and Non-spontaneous Speech which includes single-word utterances (SW).

To enhance model robustness and generalization capabilities, we augment the SAPC dataset with additional speech sources.
We incorporate the TORGO dataset \cite{torgo} (5.71 hours of natural dysarthric speech and 4.42 hours of synthetic speech) to increase exposure to diverse dysarthric speech patterns.
Additionally, we include VoxPopuli \cite{voxpopuli} (500+ hours of native and non-native English speech) to ensure the model maintains strong performance on general English speech patterns while adapting to dysarthric speech characteristics.

\subsection{Data Processing and Augmentation}
To address the specific challenges of dysarthric speech recognition, we implement a data augmentation pipeline inspired by CrisperWhisper \cite{crisper_whisper}. 
Our approach focuses on creating realistic variations in speech timing and background conditions, which are particularly relevant for dysarthric speech processing:

\begin{enumerate}
    \item Following CrisperWhisper's positional augmentation strategy, we inject background noise from MUSAN dataset \cite{musan} at either the beginning or end of each audio segment with 50\% probability, while maintaining the original transcription. 
    This approach prevents overfitting to specific temporal positions in the encoder output while simulating real-world acoustic conditions.
    \item We introduce pure noise samples from MUSAN \cite{musan} during training with 1\% probability and require empty predictions as output. 
    This technique aims to mitigate hallucinations, a common issue in generative ASR methods like Whisper.
    \item With 25\% probability, we apply either time stretching (0.85x-1.15x speed range) to address variable speaking rates in dysarthric speech, or SpecAugment \cite{park19e_interspeech} with frequency and time masking to enhance spectral robustness.
\end{enumerate}

\subsection{Transcription Format}
The SAPC dataset provides detailed transcriptions that capture both the main speech content and disfluencies.
For spontaneous speech utterances, transcriptions include the input question in square brackets, while prompted speech (i.e.,  digital assistant commands or read sentences) contains only the speaker's response.
We preserve all content in parentheses as they mark speech events with clear acoustic evidence in the recordings\footnote{Examples provided are for illustration purposes only and do not represent actual data from the SAPC dataset.}.
For instance, partial word attempts or sound repetitions are transcribed as ``(che- che-) checkout'' or ``(d-*) demo''.
Additionally, round brackets mark contextual events through special tags: ``(cs:...)'' indicates interviewer speech, while ``(ss:...)'' denotes speaker utterances not related to the main prompt.
In the target transcription for ASR we remove content within square brackets [. . .] while retaining all content within round brackets (. . .), as these segments have corresponding acoustic evidence in the speech signal.

\subsection{Model Architecture and Training}
Our framework implements a two-stage approach combining ASR and Generative Error Correction.

\vspace{1mm}
\noindent
\textbf{ASR Stage.}
We evaluate two Whisper configurations to analyze zero-shot capabilities and adaptation benefits. 
The first uses Whisper \textsc{Large-v3}\footnote{\url{https://huggingface.co/openai/whisper-large-v3}} without adaptation, while the second fine-tunes Whisper \textsc{Large-v2}\footnote{\url{https://huggingface.co/openai/whisper-large-v2}} on our combined dataset. 
We opted for Whisper \textsc{Large-v2} fine-tuning based on preliminary experiments showing that \textsc{Large-v3} exhibited significant instability when adapted to dysarthric speech.
Specifically, fine-tuning \textsc{Large-v3} led to severe degradation in transcription quality, characterized by repetitive pattern artifacts and incomplete utterances.
This suggests that \textsc{Large-v3}'s pre-trained representations, while effective for general ASR, may either be less suitable for domain adaptation to dysarthric speech patterns or require additional research to identify optimal fine-tuning strategies.
For ASR fine-tuning, we use the AdamW optimizer \cite{adamw} with a peak learning rate of $5\times10^{-6}$.
The training runs for 5 epochs using a batch size of 64.
We employ a linear learning rate schedule with 10\% warmup steps followed by linear decay.

\vspace{1mm}
\noindent
\textbf{GER Stage.}
The GER stage uses two sizes of the FlanT5 model family: XL (3B parameters) and XXL (11B parameters). 
Given the large model sizes, we employ LoRA adaptation with rank $r=16$, $\alpha=32$, targeting all linear layers. 
This introduces approximately 1\% additional trainable parameters while keeping the base model frozen.
For GER training, we use AdamW with a peak learning rate of $1\times10^{-4}$  and the same warmup-decay schedule as ASR fine-tuning.
Training runs for 10 epochs with an effective batch size of 32 on 4 NVIDIA A100 80GB GPUs, the same hardware setup used for ASR fine-tuning.

\subsection{Evaluation Protocol}
The challenge provides three evaluation sets: development, \textsc{test-1}, and \textsc{test-2}. 
The development set enables model tuning and hyperparameter optimization, while \textsc{test-1} provides ongoing feedback through a public leaderboard. 
\textsc{test-2} remains hidden until the challenge closes, ensuring unbiased final evaluation.
Performance is measured using two complementary metrics. 
Word Error Rate (WER) captures literal transcription accuracy through normalized string edit distance. 
The Semantic Score (SemScore) combines three normalized components with different weights: BERTScore \cite{BERTScore} for semantic similarity, phonetic distance for pronunciation accuracy, and natural language inference for bidirectional entailment between hypothesis and reference. 
Each utterance is evaluated against two reference transcripts: one with disfluencies and one without. 
Following challenge guidelines, the system selects the reference with the lower normalized edit distance for scoring.
This approach supports both verbatim and cleaned transcriptions, as each may be more appropriate depending on the application context.

\section{Results and Analysis}
\label{sec:results}

We evaluate our framework through a series of experiments examining both overall performance and specific components.
Our analysis focuses on understanding how acoustic and linguistic modeling capabilities contribute to dysarthric speech recognition.

\vspace{-2mm}
\subsection{Overall System Performance}

Table~\ref{tab:asr-performance} shows the performance of different system configurations across development and test sets.
The baseline Whisper \textsc{Large-v3} model achieves 11.60\% WER on the development set without any adaptation, demonstrating strong zero-shot capabilities for dysarthric speech.
Performance varies considerably across evaluation sets: this baseline achieves 11.39\% WER on \textsc{test-1} but degrades to 14.49\% WER on \textsc{test-2}, suggesting different levels of speech complexity in the test sets.
Adding generative error correction with the 11B parameter model reduces development set WER to 7.34\% and \textsc{test-1} WER to 10.63\%, showing that linguistic knowledge can effectively disambiguate ASR hypotheses.
The improvement in semantic scores (from 83.91 to 90.32 on development set) suggests that GER not only corrects transcription errors but also enhances semantic understanding.
Fine-tuning Whisper \textsc{Large-v2} provides substantial gains, achieving 7.17\% WER and 91.94 SemScore without GER in the development set, indicating effective acoustic model adaptation for dysarthric speech.
Combining fine-tuning with GER achieves the best overall performance (6.40\% WER, 92.47 SemScore on development; 12.89\% WER on \textsc{test-2}), demonstrating complementary benefits from acoustic and linguistic components.
Our experiments with model scaling revealed interesting trade-offs between performance and computational requirements.
Scaling the GER model from 3B to 11B parameters shows diminishing returns on test sets, suggesting that computational resources might be better invested in other system components.
Using a single NVIDIA A100 GPU, our fine-tuned Whisper \textsc{Large-v2} achieves an average inference time of $0.55$s per sample, while the complete pipeline including GER averages $0.69$s per sample -- remaining practical for real-world applications despite the high computational demands of the models.
The consistent performance gap between development and test sets (approximately 6 percentage points in WER) highlights the challenge of building robust systems for diverse speaking patterns.

\vspace{-1mm}
\subsection{Analysis of N-best List Generation}
\vspace{-1mm}

Table~\ref{tab:ablation_n_best} demonstrates the impact of N-best list size on recognition accuracy using Whisper \textsc{Large-v3} on the development set.
Increasing the number of hypotheses from 1 to 20 provides diminishing returns, with most gains achieved by the first 5 hypotheses (WER reduction from 12.37\% to 11.71\%).
This pattern suggests that the ASR model captures relevant acoustic information within its top hypotheses, supporting our first research question about acoustic modeling capacity.
The modest improvements beyond 5 hypotheses (only 0.11\% WER reduction from 5 to 20 hypotheses) indicate that additional candidates often represent minor variations rather than fundamentally different interpretations.
Our diversity-based selection algorithm helps maintain meaningful variation while keeping computational costs manageable.

\begin{table}[t]
\centering
\caption{Performance breakdown by utterance category on development set. Categories: DAC (Digital Assistant Commands), SN (Sentences from Novels), SS (Spontaneous Speech), SW (Single Words).}
\vspace{-2mm}
\label{tab:category-performance}
\setlength{\tabcolsep}{4pt}
\resizebox{\columnwidth}{!}{%
\begin{tabular}{@{}lc|cc|cc@{}}
\toprule
\multirow{2}{*}{Category} & \multirow{2}{*}{\#} & \multicolumn{2}{c|}{w/o GER} & \multicolumn{2}{c}{w/ GER} \\
& & WER $\downarrow$ & SemScore $\uparrow$ & WER $\downarrow$ & SemScore $\uparrow$ \\ \midrule
DAC & 15,066 & 6.47 & 92.77 & 5.61$_{\textcolor{positive}{-0.86}}$ & 93.28$_{\textcolor{positive}{+0.51}}$ \\
SN & 3,746 & 5.74 & 92.96 & 5.21$_{\textcolor{positive}{-0.53}}$ & 93.41$_{\textcolor{positive}{+0.45}}$ \\
SS & 2,313 & 10.86 & 87.27 & 10.31$_{\textcolor{positive}{-0.55}}$ & 88.14$_{\textcolor{positive}{+0.87}}$ \\
SW 
& 130 & 63.08 & 49.84 & 63.08$_{\textcolor{black}{+0.00}}$ & 49.46$_{\textcolor{negative}{-0.38}}$ \\ \bottomrule
\end{tabular}
}
\vspace{-5mm}
\end{table}

\subsection{Performance Across Utterance Types}
\label{subs:performance_utter}

Table~\ref{tab:category-performance} presents a detailed breakdown of system performance across different utterance categories in the development set.
The results reveal distinct patterns in how GER affects different speaking styles.
For digital assistant commands (DAC), which represent the majority of the dataset, GER reduces WER by 0.86\% absolute while improving semantic scores by 0.51 points.
Similar improvements appear in novel sentences (SN), with a 0.53\% WER reduction and a 0.45-point increase in semantic accuracy, suggesting GER's effectiveness on structured speech patterns.
Spontaneous speech (SS) shows comparable WER reduction (0.55\%) but achieves a larger semantic improvement (0.87 points), indicating that GER's corrections are particularly meaningful for conversational speech.

\noindent
Table~\ref{tab:ger-example} presents a case where our diversity-based hypothesis selection preserves the correct transcription.
While standard beam search ranks hypotheses by confidence score alone, our approach explicitly selects diverse alternatives.
This diversity is evident in the example, where despite errors in the top hypothesis (\textit{play}, \textit{set}, \textit{Monday}), the correct transcription with \textit{pet}, \textit{sits}, and \textit{lap} is preserved as the second hypothesis.
By analyzing these diverse candidates, the GER model successfully identifies and selects the correct interpretation.
However, single words (SW) transcriptions reveals a critical limitation, with extremely high WER (63.08\%) and no improvement from GER.
The analysis of SW errors shows that the system consistently misinterprets isolated words as short common phrases or expressions (e.g., ``football'' → ``What law?'', ``tape'' → ``Hey'') or longer phrases (e.g., ``Birmingham'' → ``Lonnie Hill''), suggesting a strong bias towards generating complete utterances rather than single words.
These findings highlight the impact of speaking style on dysarthric speech recognition, where our approach benefits both structured and conversational speech while isolated word recognition remains an open challenge.

\begin{table}[t]
\caption{A qualitative example from SS, showing diverse N-best hypotheses with correct transcription in \positive{green} and errors in \negative{red}.}
\vspace{-2mm}
\label{tab:ger-example}
\resizebox{\columnwidth}{!}{%
\begin{tabular}{@{}rl@{}}
\toprule
\textbf{Reference} & My favorite pet is the one that sits on my lap. \\
\midrule
\textbf{ASR Output} & My favorite \negative{play} is the one that's \negative{set} on \negative{Monday}. \\
\textbf{GER Output} & My favorite \positive{pet} is the one that \positive{sits} on my \positive{lap}. \\
\midrule
\multicolumn{2}{@{}l}{\textbf{N-best Hypotheses:}} \\
1. & My favorite \negative{play} is the one that's \negative{set} on \negative{Monday}. \\
2. & My favorite \positive{pet} is the one that \positive{sits} on my \positive{lap}. \\
3. & My favorite \negative{player} is the one that's in \negative{Orlando}. \\
4. & My favorite \negative{play} is the ones that sit on the. \\
5. & My favorite \negative{pick} is the one that said "\negative{Wonder}." \\
\bottomrule
\end{tabular}
}
\vspace{-5mm}
\end{table}

\vspace{-1mm}
\section{Conclusions}
\vspace{-1mm}

Our investigation through the Speech Accessibility Project Challenge reveals both opportunities and challenges in dysarthric speech recognition.
The strong performance of zero-shot Whisper (11.60\% WER) answers our first research question, demonstrating that general-purpose ASR models can effectively capture dysarthric speech patterns.
The consistent improvements from GER across speaking styles addresses our second question, validating language models' ability to leverage multiple hypotheses, though with diminishing returns at larger scales.
However, important limitations emerged: the poor performance on isolated words (63.08\% WER) suggests the evaluated models may overly favor complete utterances, while the significant gap between development and test set performance indicates generalization challenges across different dysarthric speech patterns.
These findings point to two promising research directions: investigating specialized architectures for isolated word recognition, and developing robust adaptation techniques to better handle the variability of dysarthric speech patterns.

\section{Acknowledgements}
This work has been partially supported by the "D.A.R.E. – Digital Lifelong Prevention" project (code: PNC0000002, CUP: B53C22006450001), co-funded by the Italian Complementary National Plan PNC-I.1 Research initiatives for innovative technologies and pathways in the health and welfare sector (D.D. 931 of 06/06/2022) and by the European Union – Next Generation EU under the National Recovery and Resilience Plan (PNRR) – M4 C2, Investment 1.1: Fondo per il Programma Nazionale di Ricerca e Progetti di Rilevante Interesse Nazionale (PRIN) - PRIN 2022 - "SHAPE-AD" (CUP: J53D23007240008).

\bibliographystyle{IEEEtran}
\bibliography{mybib}

\end{document}